\documentclass[10pt,twocolumn,letterpaper]{article}

\usepackage{wacv}              

\usepackage{appendix}
\usepackage{graphicx}
\usepackage{booktabs}
\usepackage{multirow}
\usepackage{amsmath}
\usepackage{amssymb}
\usepackage{comment}
\usepackage{subcaption}
\usepackage[accsupp]{axessibility}  %

\usepackage[pagebackref,breaklinks,colorlinks,citecolor=blue]{hyperref}

\usepackage[capitalize]{cleveref}
\crefname{section}{Sec.}{Secs.}
\Crefname{section}{Section}{Sections}
\Crefname{table}{Table}{Tables}
\crefname{table}{Tab.}{Tabs.}

\def\wacvPaperID{449} 
\def\confName{WACV}
\def\confYear{2025}

\begin{document}

\title{Realistic and Efficient Face Swapping: A Unified Approach with Diffusion Models}


\author{Sanoojan Baliah$^{1}$ \quad  Qinliang Lin$^{2}$ \quad Shengcai Liao$^{3}$ \quad Xiaodan Liang$^{1}$ \quad Muhammad Haris Khan$^{1}$\\
$^1$MBZUAI, UAE \quad $^2$Shenzhen University, China \quad $^3$Core42, UAE\\
{\tt\small \{sanoojan.baliah, muhammad.haris\}@mbzuai.ac.ae}
}
\maketitle

 \begin{figure*} [!htb]
     \centering
     \includegraphics[width=\linewidth]{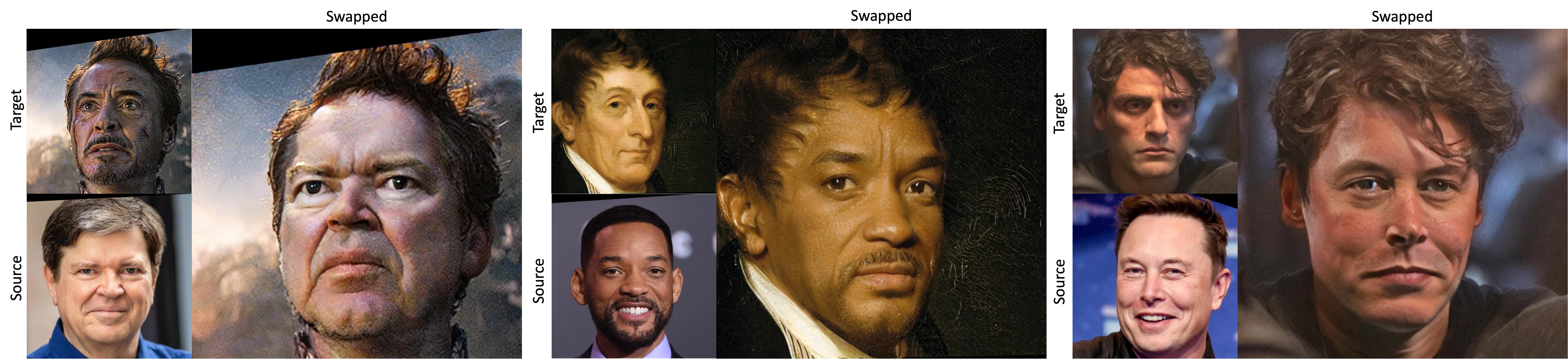}
     \caption{We show the Face Swapped results of our method. In all groups, the swapped face(right) should contain the identity information from source (left-bottom) and pose, expression and lighting conditions from target (left-top). All the images are in $512 \times 512$ resolution.}
     \label{fig:teaser_fig}
 \end{figure*}
\begin{abstract}
Despite promising progress in face swapping task, realistic swapped images remain elusive, often marred by artifacts, particularly in scenarios involving high pose variation, color differences, and occlusion. To address these issues, we propose a novel approach that better harnesses diffusion models for face-swapping by making following core contributions. (a) We propose to re-frame the face-swapping task as a self-supervised, train-time inpainting problem, enhancing the identity transfer while blending with the target image. (b) We introduce a multi-step Denoising Diffusion Implicit Model (DDIM) sampling during training, reinforcing identity and perceptual similarities. (c) Third, we introduce CLIP feature disentanglement to extract pose, expression, and lighting information from the target image, improving fidelity. (d) Further, we introduce a mask shuffling technique during inpainting training, which allows us to create a so-called universal model for swapping, with an additional feature of head swapping. Ours can swap hair and even accessories, beyond traditional face swapping. 
Unlike prior works reliant on multiple off-the-shelf models, ours is a relatively unified approach and so it is resilient to errors in other off-the-shelf models. 
Extensive experiments on FFHQ and CelebA datasets validate the efficacy and robustness of our approach, showcasing high-fidelity, realistic face-swapping with minimal inference time. Our code is available at \href{https://github.com/Sanoojan/REFace}{REFace}.
\end{abstract}

\section{Introduction}
\label{sec:intro}

Face-swapping is a fascinating yet challenging problem in the field of computer vision, aiming to seamlessly transfer the identity features of a source image onto a different target image while preserving the target's pose, expression and background \cite{chen2020simswap,DeepFakes,8578800}. 
%
The reliance on GANs for face-swapping has led to notable breakthroughs \cite{li2019faceshifter,nirkin2019fsgan,chen2020simswap,wang2021hififace}, where the identity features from a source face are extracted and fused with the target face's attribute features. However, 
training GANs requires laborious hyperparameter search especially for face-swapping where multiple off-the-shelf models\cite{li2019faceshifter,nirkin2019fsgan} are included and such methods often suffer from the mode collapse problem \cite{thanh2020catastrophic}. Also, GAN-based method are prone to producing artifacts, especially under large pose variations, color differences, and occlusions \cite{kammoun2022generative}. These challenges necessitate a closer examination of alternative approaches.

In recent years, the Diffusion model \cite{ramesh2022hierarchical,rombach2022high} has emerged as a powerful contender in image generation tasks, demonstrating success in providing stable training and favorable outcomes in diversity and fidelity. Such stability and desirable performance of the Diffusion model makes it a compelling choice for potentially addressing the inherent difficulties in face-swapping. Some pioneering works\cite{kim2022diffface,zhao2023diffswap} attempt to apply diffusion modelling to the task of face-swapping with some success. Diffswap \cite{zhao2023diffswap} initially capitalizes on the diffusion model by reframing face-swapping as inference-time conditional inpainting. It is then supplemented with the midpoint estimation techniques and 3D-aware masked diffusion to obtain a high fidelity swapped face. However, this comes at the cost of more computationally intensive denoising steps in inference, rendering it a resource-expensive method. As for DiffFace \cite{kim2022diffface}, it injects ID embedding as condition into the diffusion model and imposes various face guidance constraints, however again at the inference stage, and so presents itself with the following two limitations: (1) The inference process becomes time-consuming due to the gradient computation performed during testing, leading to significant time overhead.~(2) Swapped faces often produce noise artifacts and contain source image accessories if present. Both diffusion-based methods employed in face-swapping predominantly accomplish the swapping process during the inference stage, thereby contributing to prolonged computational time. Further, these methods encounter challenges in effectively adapting to variations in the background lighting conditions. 

Towards overcoming the existing issues, this paper proposes a novel approach for face-swapping that aims to better harness the potential of the Diffusion model. \textbf{(1)} We frame the face-swapping problem as a self-supervised, train-time inpainting task, where the Diffusion model learns to transfer the identity features seamlessly while blending with the target image. \textbf{(2)} To reinforce this process, we introduce an N-step Denoising Diffusion Implicit Model (DDIM) sampling during training to enforce ID similarity and perceptual similarity at each step. This enhances the model’s performance, particularly in ID transferability. Additionally, this approach allows our method to achieve efficient, minimal-step inference. \textbf{(3)} Next, we leverage the CLIP feature to enhance the realism as well as expression maintenance with the target image by the disentanglement of CLIP feature. \textbf{(4)} Apart from face-swapping, our model can perform more challenging swaps, including head swapping. In head swapping, we replace the entire head, including hair, while preserving the identity from the source image and maintaining the pose and expression of the target image with high fidelity. To achieve this we propose a mask shuffling technique in the training pipeline. Overall our approach brings a diverse and a realistic swapping model with high fidelity with minimal inference time techniques i.e., the naive DDIM inferencing. 
Also, many prior works depend on several other off-the-shelf sophisticated models such as the reenactment networks \cite{nirkin2019fsgan,zhang2019faceswapnet} and face parsing networks to obtain the detailed segmentation. Whereas our approach is simple as it needs just the overall mask region, ID feature, CLIP feature and landmarks to provides a realistic swapping even if there is some error from other models. \textbf{(5)} Through extensive experimentation on both the FFHQ and CelebA datasets, we substantiate the superior robustness of our models with minimal inference time across varied datasets and scenarios (see Fig.~\ref{fig:teaser_fig}).  




\section{Related works}
\label{sec:Related works}


\noindent \textbf{GAN-based methods:} 
GANs are a powerful tool for performing face-swapping by extracting identity features from the source face and seamlessly fusing them with the target face attribute features, resulting in high-fidelity swapped faces. To avoid the subject-specific training, FSGAN\cite{nirkin2019fsgan} has been proposed to be a subject-agnostic method that incorporates a reenactment network and an inpainting network. By leveraging face landmarks and segmentation, the method aims to recreate the source image in alignment with the target image. Unlike FSGAN, HifiFace\cite{wang2021hififace} adopted other prior structural information like the 3D shape-aware Identity feature to guide the high quality swapped face. However, structural prior-guide models are usually limited to imprecise face prior information. In contrast to above two methods, SimSwap\cite{chen2020simswap} and Faceshifter\cite{li2019faceshifter} are general and prior information-free methods. Faceshifter proposed a two-stage framework to realize high fidelity and occlusion aware face-swapping, while SimSwap proposed ID injection Module and Weak Feature Matching Loss to help the network possess a good attribute preservation ability. 
A recent breakthrough in face-swapping, employing StyleGAN~\cite{karras2019style,karras2020analyzing}, has yielded an effective solution for achieving superior quality face swaps at high resolutions \cite{zhu2021one, xu2022high, gao2021information, xu2022region,liu2023fine}.
For instance, MegaFS\cite{zhu2021one} first projected source and target image into hierarchical representation space 
and then fed them into StyleGAN generator to obtain the swapped face. 
E4s \cite{liu2023fine} proposes a region GAN inversion on the explicit disentanglement of texture and shape features.
%
%
%

%

\noindent \textbf{Diffusion-based methods:}
In contrast to GANs, diffusion models offer a more stable training approach and demonstrate desirable outcomes in terms of both diversity and fidelity. Intuitively, utilizing the diffusion model for the face-swapping task can also serve as a strong comparative baseline. DiffFace\cite{kim2022diffface} made the first attempt to employ a diffusion-based framework that aims to achieve high-fidelity and controllable face swapping. However as they use ID features only to train the model and swapping is entirely done at the inference stage, this approach significantly burdens the inference process, leading to low throughput even at lower resolutions. Additionally, it often tends to produce noise artifacts, particularly in the eye region.


%
Another diffusion model-based work, namely DiffSwap\cite{zhao2023diffswap} utilized a powerful diffusion model to redefine face-swapping as inference time conditional inpainting. They also introduced a 3D-aware masked diffusion approach at inference to ensure the explicit consistency of facial shape. Although Diffswap trains the diffusion model using combined conditions from source and target images, its loss formulation compels the model only to reenact the source image using the target image’s landmarks. However, an essential blending step is performed through masked fusion during the inference stage, which is inefficient. Additionally, the method struggles to achieve effective ID transferability. To address these issues, we train the diffusion model as a conditional inpainting network, which fully mimics face swapping during the training stage. We also supplement the conditioning with powerful CLIP features, using our proposed disentanglement strategy to improve ID transferability while maintaining the target’s pose, expression, and lighting. Our approach simplifies the inference process, allowing our method to produce high-quality face swaps with significantly fewer inference steps and less inference time compared to DiffFace\cite{kim2022diffface} and Diffswap\cite{zhao2023diffswap}.

\section{Methodology}
\label{sec:Methodlolgy}

\subsection{Preliminaries}

\noindent\textbf{Problem setting:} In face-swapping, we formally define a target image $x^{tar}$ and a source image $x^{src}$ where the source image's identity features should be swapped to the target image's face region while preserving target's pose, expression, lighting condition and background \cite{chen2020simswap}. Here, we swap a face region $m^{tar}$ of the target with the information from the source image's face region $m^{src}$ and produce the swapped image $x^{swap}$. Apart from preserving the above mentioned properties, a practical swapped face should be plausible and properly merge to the target image's outer region ($1-m^{tar}$) without producing any artifacts. Furthermore, during training, we use the term reference image $x^{ref}$, which serves as a source $x^{src}$ image during inference. 

\begin{figure*}
     \centering
     \includegraphics[width=\linewidth]{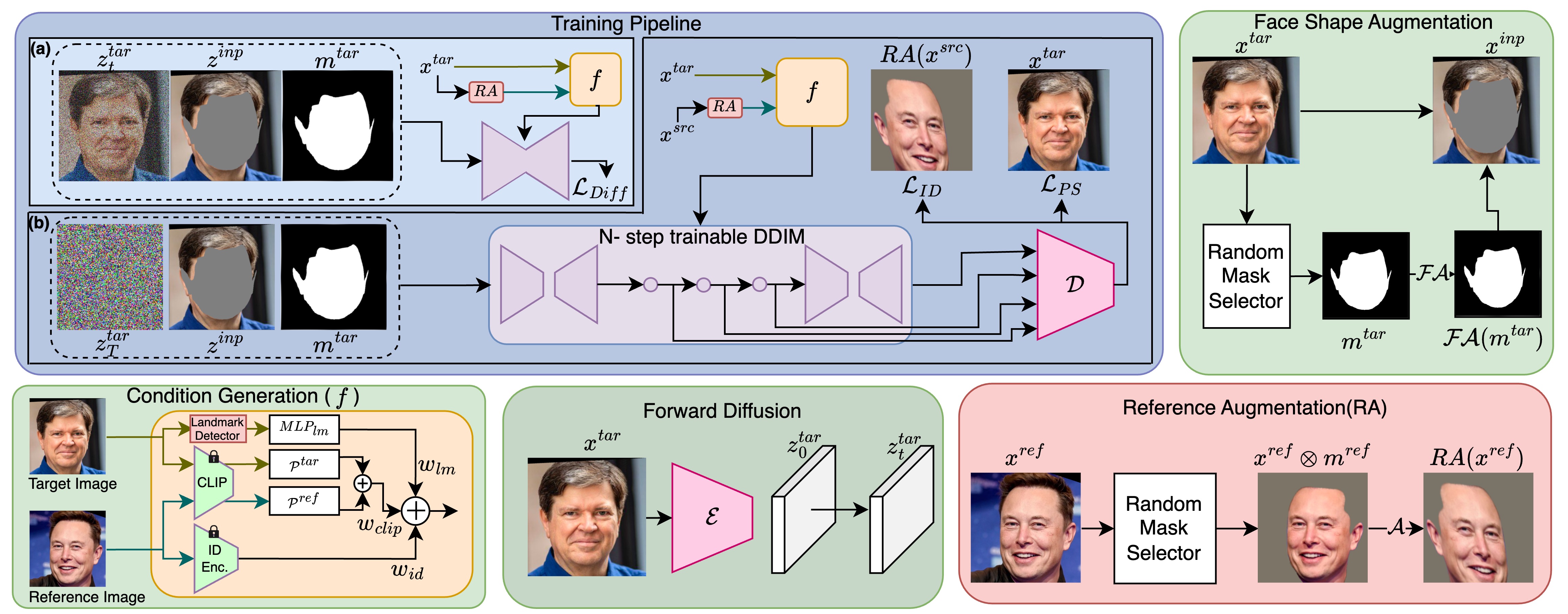}
     \vspace{-1em}
     \caption{Our architecture pipeline for face swapping. The training pipeline contains two major components (a) the inpainting diffusion training, where we utilize the same image as reference (with reference augment) and inpaint (with face shape augment). (b) Here we provide different images and enforce identity loss and perceptual similarity with source and target respectively. On the bottom left, we depict the condition generation. Further, all the $z$-s in training pipeline are latents obtained from the forward diffusion, but for better understandability we show as images.}
     \vspace{-2em}\label{fig:architecture}
 \end{figure*}

\noindent\textbf{Diffusion model:} 
Typically, denoising in diffusion models \cite{sohl2015deep} is executed using a U-Net architecture, trained to anticipate noise patterns introduced during the diffusion forward process at any given timestep $t$. A diffusion model can be defined as a $T$-step forward process and a $T$-step reverse process. To more formally introduce let us define the diffusion forward process $z_t$ at $t^{th}$ timestep is obtained from $z_{t-1}$ by adding a gaussian noise defined as $q(z_t|z_{t-1}) \sim \mathcal{N}(\sqrt{\alpha_t} z_{t-1}, (1-\alpha_t)I)$. Here $\{\alpha_t\}_{t=1}^T$ represents a pre-defined coefficient sequence regulating the noise variance schedule.  Alternatively, the forward process for any time step $t$ from the initial latent distribution $z_0$ can be represented by a closed-form solution,
\begin{equation}
     z_t=\sqrt{\Bar{\alpha_t}}z_0 + \sqrt{1-\Bar{\alpha_t}}\epsilon.
     \label{Eq.Forward_diff}
\end{equation}

  Where, $\Bar{\alpha_t}=\prod_{i=1}^t \alpha_i$ and $\epsilon$ is sampled from gaussian normal distribution. Here the diffusion model is a learned denoising network $\epsilon_{\theta}$, trained to predict the $\epsilon$,

\begin{equation}
    \epsilon_{\theta}(z_t,t) \approx \epsilon=\frac{z_t-\sqrt{\bar{\alpha_t}}z_0}{\sqrt{1-\Bar{\alpha_t}}} 
\end{equation}

In the vanilla diffusion model, the training is done to predict the noise added in any arbitrary random timestep chosen from a uniform distribution, that is, $t \sim \mathcal{U}(1,T)$, where $ T$ is the maximum timestep in the diffusion process that is sufficient enough to make any image a complete Gaussian noise for the given noise schedule. The training objective is often decomposed as the stepwise KL-divergence between the predicted distribution and the posterior distribution. This can be done within the noise or the noised latents. However, most of the works on diffusion do so within the noise distributions. This objective function can be further simplified as:

\begin{equation}
    \mathcal{L}_{DDPM}=\mathbb{E}_{t,z_0,\epsilon}\parallel \epsilon_\theta (z_t,t) - \epsilon \parallel _2^2.\\
\end{equation} 

Once the diffusion model $\epsilon_{\theta}$ is trained, the denoising stage can be modeled in different ways \cite{song2022denoising,liu2022pseudo}. \cite{song2022denoising} proposed Denoising Diffusion Implicit Models (DDIM) to compute the predicted clean data point which helps to make the denoising stage faster. 
\begin{equation}
    \label{eq:ddim_pred}
  \hat{z}_0(z_t,t)=\frac{z_t-(\sqrt{1-\Bar{\alpha}_t})\epsilon_\theta(z_t,t)}{\sqrt{\Bar{\alpha_t}}}
\end{equation}

Further, the mapping between the latent distribution is done using the perceptual compression auto encoder  model to reduce the computation cost yet giving the handling capacity of high resolution images. As in Fig~\ref{fig:architecture} (Forward diffusion), given an image $x\in \mathbb{R}^{H \times W \times 3}$, $\mathcal{E}$ is an encoder that encodes $x$ into a low resolution latent representation $z=\mathcal{E}(x)$, and the decoder $\mathcal{D}$ reconstructs the image from the latent $z$, giving $\Tilde{x}=\mathcal{D}(z)$ where $z\in \mathbb{R}^{h \times w \times c}$ \cite{rombach2022highresolution}. 



\begin{figure}
    \centering
    \includegraphics[width=\linewidth]{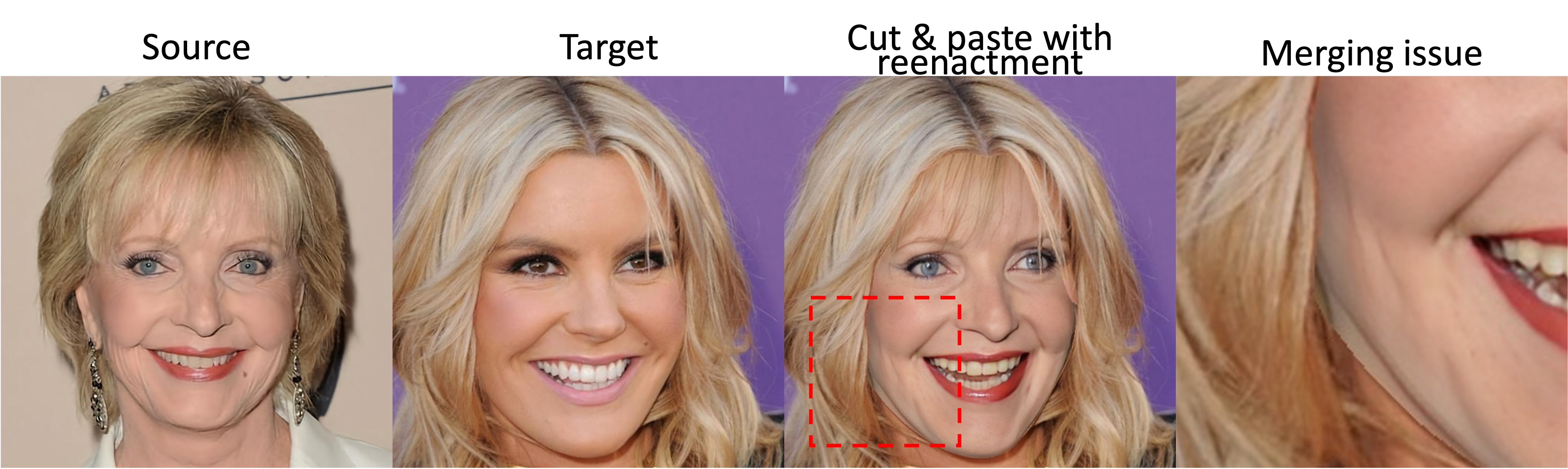}
    \caption{Example reenactment cut and paste swapping. Zoom in and view at the boundaries of outcome which has merging issues.}
    \vspace{-2em}\label{fig:reenactment_example}
\end{figure}

\noindent\textbf{Naive solution:}
The cut-and-paste approach for face-swapping, where the face from the source image is cropped and affixed onto the target image background, often fails to align the pose of the source face with that of the target.
An alternative approach leverages reenactment networks \cite{nirkin2019fsgan} to adjust the pose and orientation of the face of the source image before applying the cut-and-paste technique. However, the resulting images may still exhibit artificial artifacts, particularly at the merging boundaries (Fig.~\ref{fig:reenactment_example}).

Previous studies have tackled face-swapping through two primary avenues \cite{chen2020simswap}: (1) altering the target face image based on the identity features of the source image. However, this method often falls short in effectively transferring identity features, and (2) reconstructing the swapped image using identity features as a blueprint, then integrating it into the target background. The latter approach is more promising, as it can infuse the swapped face with more authentic identity-related attributes. However, it tends to introduce more artifacts, primarily due to the merging process. Additionally, achieving accurate lighting conditions from the original image or adapting them to fit the background proves challenging, as no background information is typically utilized in the reconstruction process.

Therefore, we train the diffusion model as a conditional inpainting network, unlike other diffusion-based face-swapping approaches that train diffusion as a reconstruction or reenactment network. For instance, DiffFace trains the model as a reconstruction network using the ID features as a condition. Our approach is primarily motivated by exemplar-based image editing as demonstrated in \cite{yang2022paint}. This technique allows for image editing using CLIP image feature conditions from an example image, altering the context of a picture to match the given example image within the masked inpaint region. As this method produces photorealistic image inpainting, it serves as a robust baseline for face-swapping tasks.
However, while \cite{yang2022paint} performs well with natural images, it fails to preserve the pose and expression of the target image and the identity features of the source image when applied naively to face swapping. Our main goal is to achieve photorealistic face-swapping—where many other methods fall short—while maximizing identity transferability and preserving attributes such as the target’s pose, expression, and lighting.
In Section \ref{Conditional inpaint self sup}, we describe how we adapt \cite{yang2022paint} for face-swapping as conditional inpainting training. To improve ID transferability, we use a multistep trainable DDIM to enforce a multistep swapping enhancement loss, as detailed in Section \ref{subsec:multistep swapping enhancement}.
Moreover, we introduce mask shuffling in Section \ref{subsec:mask shuffling}, which enhances the versatility of our model for various swapping applications, such as head swaps. 
See Fig.~\ref{fig:architecture} for the overall architecture of our face-swapping pipeline.

\subsection{Conditional Inpainting Diffusion training}
\label{Conditional inpaint self sup}

In this section, we present our approach to conditional inpainting diffusion training, a crucial component of our face-swapping pipeline (see Fig.~\ref{fig:architecture}-Training pipeline(a)). We train the diffusion model to fill an inpainted image within a specified masked region, leveraging the features of the augmented version of the original image to guide the reconstruction process. To create this inpaint image we use our face shape augmentation. \textbf{Face Shape Augmentation ($\mathcal{FA}$)}: involves a mild shift and transformation of the target face mask (see Fig.~\ref{fig:architecture}-Face Shape Augmentation). This prevents the model from learning to trivially paste the reconstructed target image from $x^{tar}$. We find that face shape augmentation results in more robust face swapping. More details about the face shape augmentation are in appendix.

To elaborate further, for a target face image $x^{tar}$, we find its latent image $z^{tar}_0=\mathcal{E}(x^{tar})$. For any randomly chosen timestep $t$, the noised latent $z^{tar}_t(z^{tar}_0,t,\epsilon)$ is obtained using Eq.\ref{Eq.Forward_diff} (see Fig.\ref{fig:architecture}-Forward Diffusion). Additionally, the in-painting image latent $z^{inp}$ is obtained from: 

\begin{equation}
    z^{inp}=\mathcal{E}(x^{tar}\otimes \mathcal{FA}(1-{m}^{tar}))
\end{equation}
 where, $m^{tar}$ is the facial mask region that needs to be inpainted by the diffusion model, and $\mathcal{FA}$ is a face shape augmentation.
 We further compute the conditional feature as a weighted average of the CLIP\cite{radford2021learning}, Arcface\cite{deng2019arcface} ID features, and facial landmarks. Specifically, we find the reference image $x^{ref}=\mathcal{A}(x^{tar}\otimes m^{tar})$, which is the facial region, that serves as the source image at inference time. Here $\mathcal{A}$ is an augment operator comprising of random resize, horizontal flip, rotate, blur and elastic transform operations. 
 
\noindent \textbf{Condition Generation:} is a pivotal aspect of training conditional diffusion models, as the quality of the generated image heavily depends on the quality of the condition feature. We find that using only the ID features of $x^{ref}$ and the landmark features of $x^{tar}$ is insufficient for both ID feature transfer and pose preservation. To this end, we propose to leverage the CLIP\cite{radford2021learning} image feature to extract the pose and expression information from the target image, and identity information complementary to the arcface ID features from the reference. We train two different linear projections to extract such different information by disentangling CLIP feature in the same pipeline, those are: a reference projector $\mathcal{P}^{ref}$ and a target projector $\mathcal{P}^{tar}$. For the augmented reference image $RA(x^{ref})$, we obtain the CLIP image encoder feature $f^{ref}_{clip} \in \mathbb{R}^D$, Arcface ID feature $f_{id} = MLP_{id}(f^{'}_{id}(RA(x^{ref})))$ where $f_{id} \in \mathbb{R}^D$, $f^{'}_{id}$ is the Arcface featurizer and the $MLP_{id}$ is a simple linear layer. From the target image, we obtain the CLIP image encoder feature $f^{tar}_{clip}$ and facial landmark positions $l^{tgt}=f{'}_{lm} \in \mathbb{R}^{68 \times 2}$ using DLib\cite{king2009dlib}, which is further transformed as $f^{lm}=MLP_{lm} \circ l^{tgt} \in \mathbb{R}^D$. These features are then combined to create the conditioning feature $f$,
\begin{align}
    &f_{clip}= \mathcal{P}^{ref} \circ f^{ref}_{clip} + \mathcal{P}^{tar} \circ f^{tar}_{clip}\\
    &f=w_{clip} f_{clip}+w_{id} f_{id} +w_{lm} f_{lm}
\end{align}
    
where, $w_{clip},w_{id}$ and $w_{lm}$ are the weights to average the features. We empirically find that other forms of combining these, such as concatenating and stacking features, provide inferior results in the same setting (i.e. the number of epochs). The conditional feature is used as the key in each cross-attention layer in the diffusion U-Net. This stage works as a self-supervision where the diffusion model takes the input as $\{z_t, z^{inp},m^{tar}\}$, and using the condition $f$, outputs the  $\epsilon_{\theta}$ the predicted noise incurred in the diffusion pipeline from step ${t-1}$ to $t$. Thus, the loss for this stage is formulated as,

 \begin{equation}
     \mathcal{L}_{Diff}=\mathbb{E}_{t,z_0,\epsilon} \parallel 
 \epsilon_\theta (z_t,z^{inp},m^{tar},f,t) - \epsilon \parallel _2^2
 \end{equation}


\subsection{Multistep swapping enhancement loss}
\label{subsec:multistep swapping enhancement}
Though the above-mentioned (sec.~\ref{Conditional inpaint self sup}) inpainting training pipeline can achieve realistic face-swapping compared to most of the other works, this lacks the ID feature transferability. To further improve the ID feature transferability and to train the projections $\mathcal{P}^{ref}$ and $\mathcal{P}^{tar}$  which disentangles the CLIP feature, we need to enforce the swapped image $x^{swap}$'s ID similarity with $ x^{ref}(=x^{src}) $ as well as perceptual similarity with $x^{tar}$. Note that this part uses different \{$x^{tar},x^{ref}$\} pair unlike in \ref{Conditional inpaint self sup} (See Fig.~\ref{fig:architecture}-training pipeline \textbf{(a)}, and \textbf{(b)}).  We incorporate ID loss between the masked source image and the masked swapped image. However, the diffusion training pipeline is originally designed such that it predicts only the noise incurred from any timestep to the next timestep i.e., it can predict a partly denoised image for only one step and to compare the ID features this is insufficient. To this end, we propose to use the differentiable DDIM sampler at training time but with only N steps to restrict memory usage. We estimate the initial sample from each N step and compute the ID loss and perceptual similarity loss in each step using the following equaions.

\begin{align}
    \mathcal{L}_{ID}&=\sum_{i=1}^N (1-<\mathcal{D}(\hat{z}_{0,t_i}) \otimes m^{tar},x^{src} \otimes m^{src}>)\\
    \mathcal{L}_{PS}&=\sum_{i=1}^N \mathcal{L}_{PIPS}(\mathcal{D}(\hat{z}_{0,t_i}),x^{tar})
\end{align}

Where the $\hat{z}_{0,t_i}$ is obtained using eq.~\ref{eq:ddim_pred}.
Further the $t_i$'s are chosen from the predefined step schedule with N same intervals. The total loss is formulated as $\mathcal{L}_{Total}=\mathcal{L}_{Diff}+w^{\prime}_{ID} \times \mathcal{L}_{ID} +w^{\prime}_{PS} \times \mathcal{L}_{PS} $.  

\subsection{Mask shuffling}
\label{subsec:mask shuffling}
We introduce a mask shuffling mechanism into our training pipeline, enhancing the versatility of our approach. Among the 17 mask categories delineating facial regions, such as left eye, right eye, skin, background, etc., we randomly select $N_m$ masks to serve as both $m^{ref}$ and $m^{tar}$. This strategy enables our diffusion model to generalize effectively, facilitating modifications beyond facial features. Moreover, leveraging CLIP features in conjunction with identity (ID) features proves advantageous, as ID features alone may struggle to extract pertinent information from the background. This straightforward yet potent technique not only enables face-swapping but also extends to other swapping scenarios under a unified model architecture. Mask shuffling is applied during training pipline and allows for modifications beyond the facial features. It has been observed that without mask shuffling, especially in the hair region, the generated hairstyles often fail to align with the target and may appear unrealistic during head swapping.





\begin{figure*} [!tb]
     \centering
    \includegraphics[width=\linewidth]{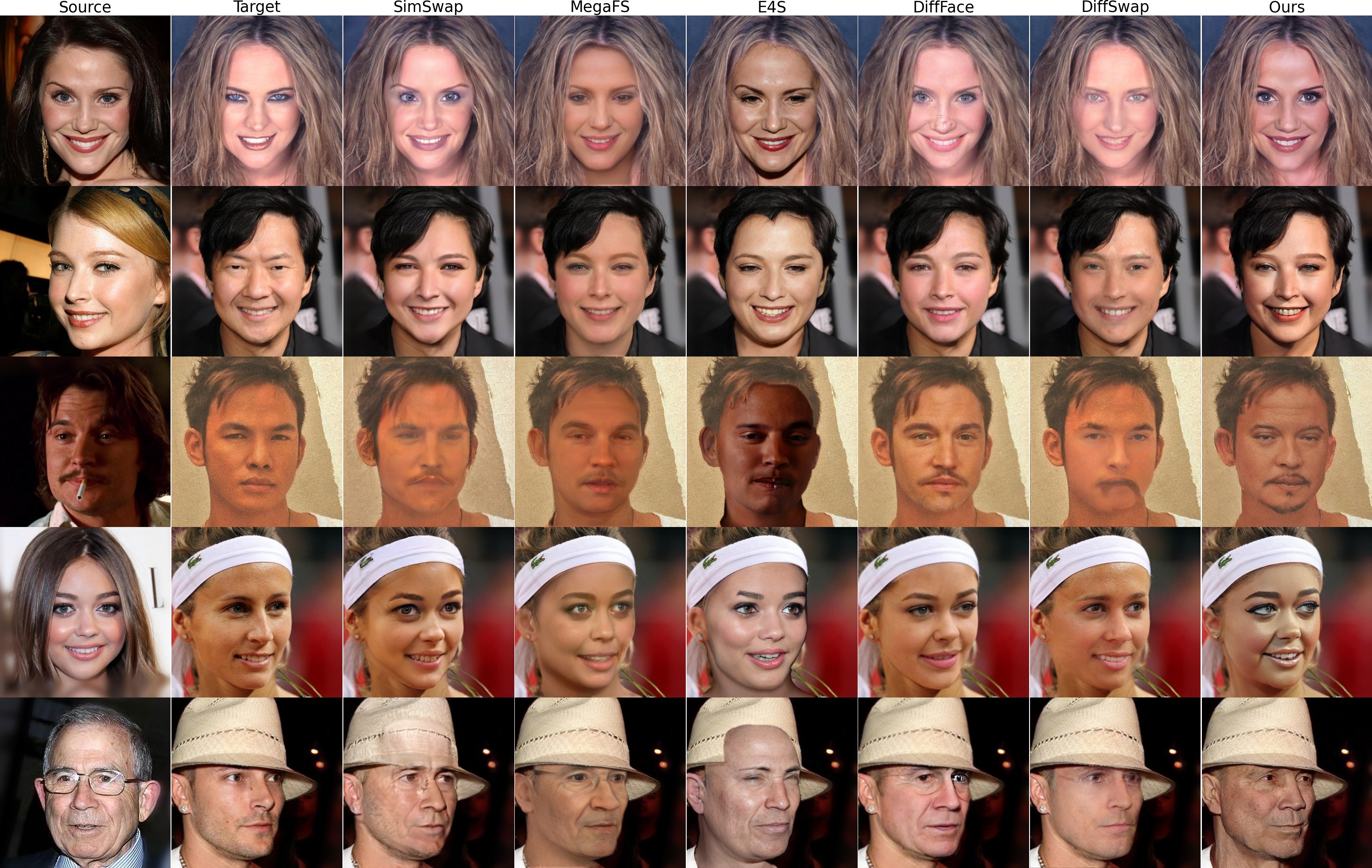}
     \caption{Qualitative comparison on CelebA dataset. Our approach demonstrates robustness in maintaining target environmental conditions (such as lighting) and preserving source ID information. Best viewed in color and zoom in.}
     \label{fig:CelebA}
     \vspace{-1em}
 \end{figure*}
\section{Experiments and Results}
\label{sec:Experiments and results}

\noindent \textbf{Datasets:}
Following prior works~\cite{liu2023fine,xu2022high} we utilize CelebAMask-HQ\cite{lee2020maskgan} as our training dataset. CelebAMask-HQ comprises 30k high-quality facial images with a resolution of $1024 \times 1024$. To avoid excessive consumption of training resources, we resize the image to 512. We use 28k samples from CelebAMask-HQ to train our model and use 2k for evaluation. We use FFHQ\cite{karras2019style} dataset for further evaluation which offers 70K $512 \times 512$ images. 

\noindent \textbf{Implementation details:} We use PyTorch to implement our framework. We train our model on 4 NVIDIA A100 GPUs(40GB). In training, we set the global batch size to 4 and initialize the learning rate as $10^{-5}$ from the stable diffusion checkpoint \cite{rombach2022high}. We train the model for 20 epochs. The weights of ID loss and LPIPS loss are 0.3, and 0.1, respectively. We use a pre-trained face recognition model\cite{deng2019arcface} to extract the face embedding, and CLIP L/14 \cite{radford2021learning} to obtain the CLIP image features. The reported qualitative and quantitaive results are produced using 50 DDIM steps.



\begin{figure*} [!htp]
     \centering
     \includegraphics[width=\linewidth]{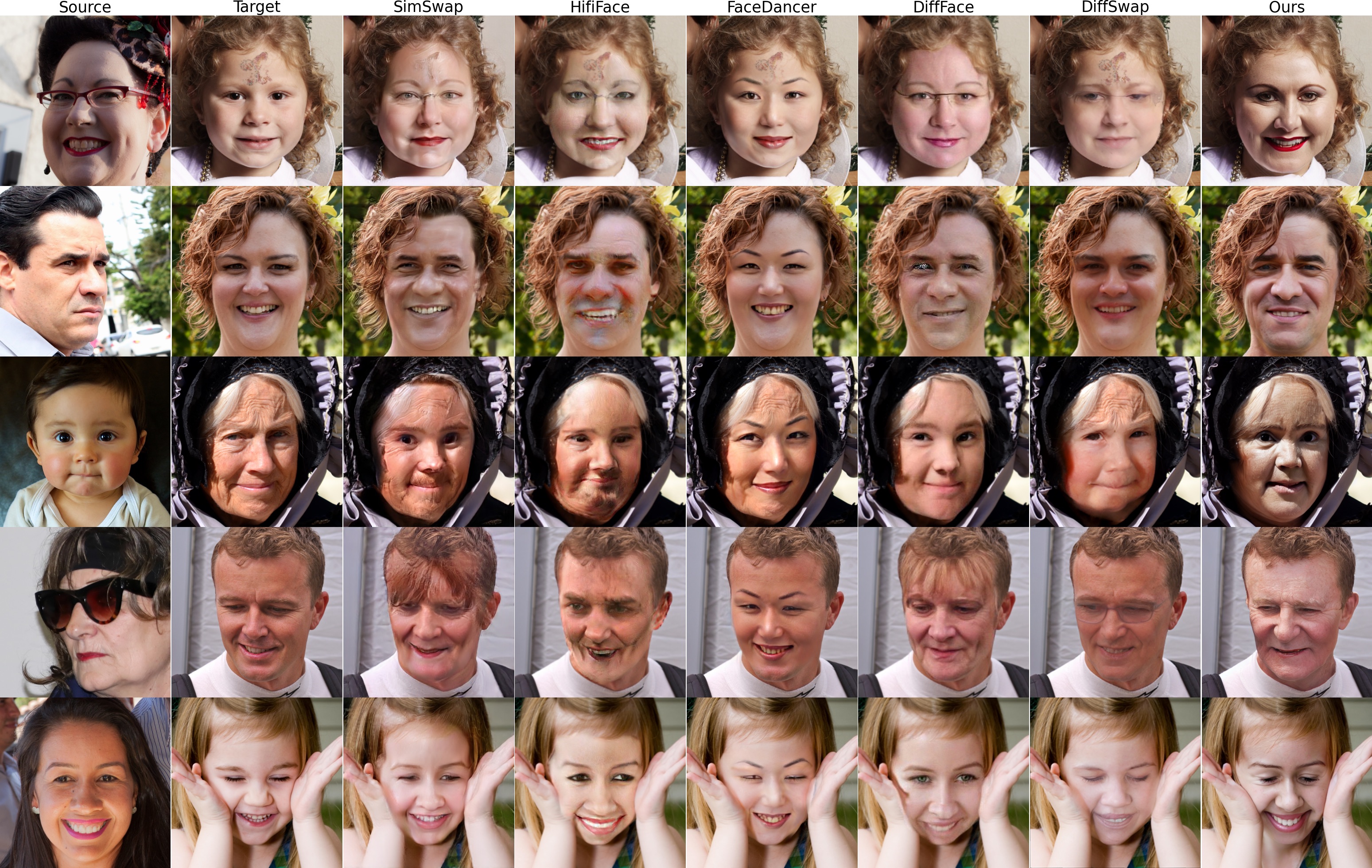}
     \caption{Qualitative comparisons of our method with SOTA face swapping methods on FFHQ dataset. Best viewed in color and zoom-in.}
     \label{fig:FFHQ} \vspace{-1em}
 \end{figure*}

\noindent \textbf{Qualitative Results:} Fig.~\ref{fig:CelebA} shows that on CelebA dataset, our method achieves more realistic and high-fidelity swapped results. Likewise, Fig.~\ref{fig:FFHQ}  demonstrates that in FFHQ dataset, our method excels in preserving both the identity and facial details of the source image, highlighting its strong generalization capabilities to open data set. Due to the combination of CLIP’s identity features and identity encoder’s features \cite{deng2019arcface}, our method can yield high-fidelity swapped face, especially on individual facial features, as well as global features including lighting, pose, and overall facial expression appearance. Furthermore, our method’s output is largely free of artifacts, especially at the edges of the faces, and it avoids generating source image accessories like eyeglasses, unlike other approaches.

\begin{figure*}[!htp]
    \centering
    \begin{subfigure}[b]{0.2835\linewidth}
        \centering
        \includegraphics[width=\linewidth]{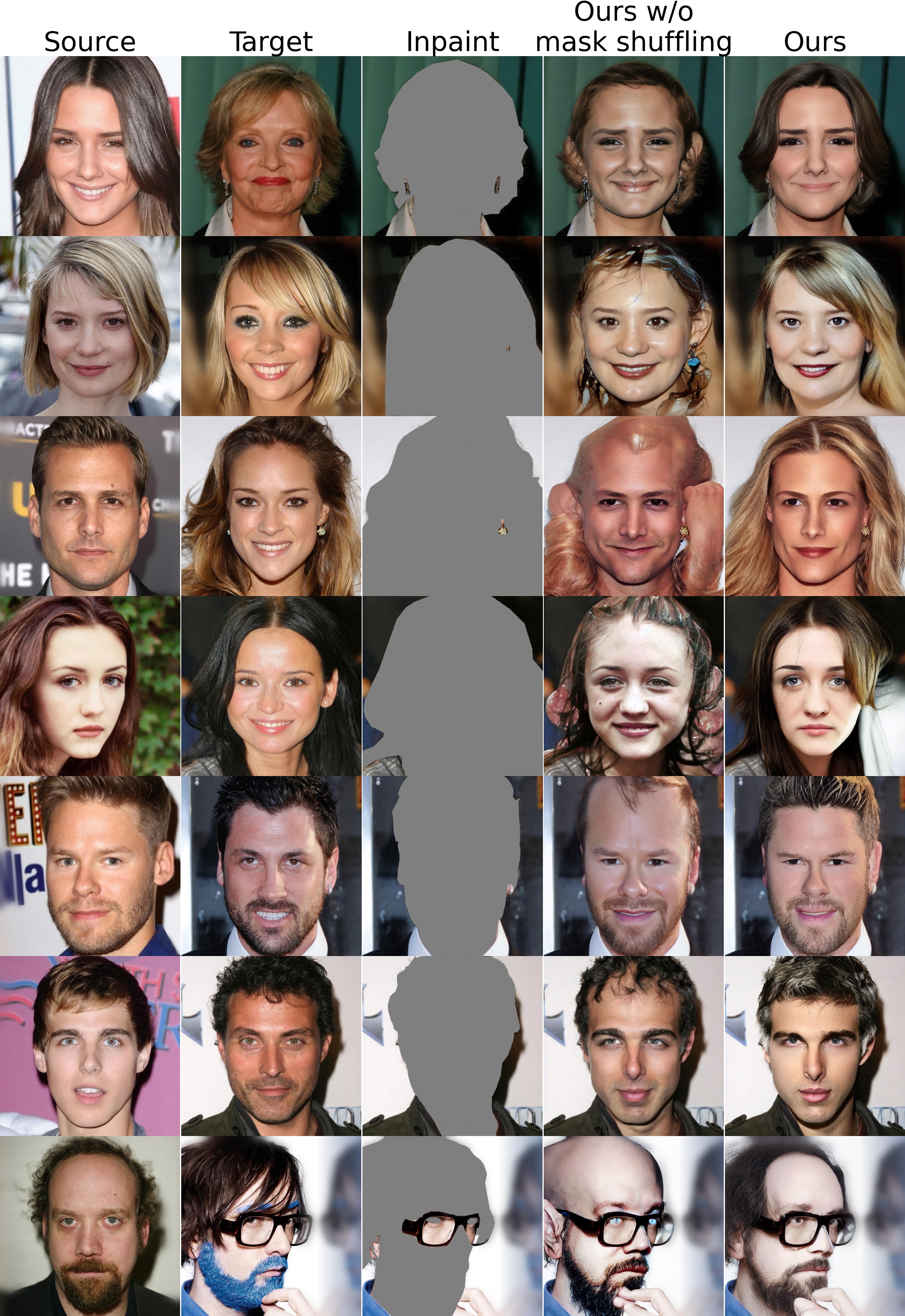}
        \caption{Results on head swapping. With the mask shuffling, our method produces robust outcome to head swap.}
        \label{fig:hair_swap_img}
    \end{subfigure}
    \begin{subfigure}[b]{0.357\linewidth}
        \centering
        \includegraphics[width=\linewidth]{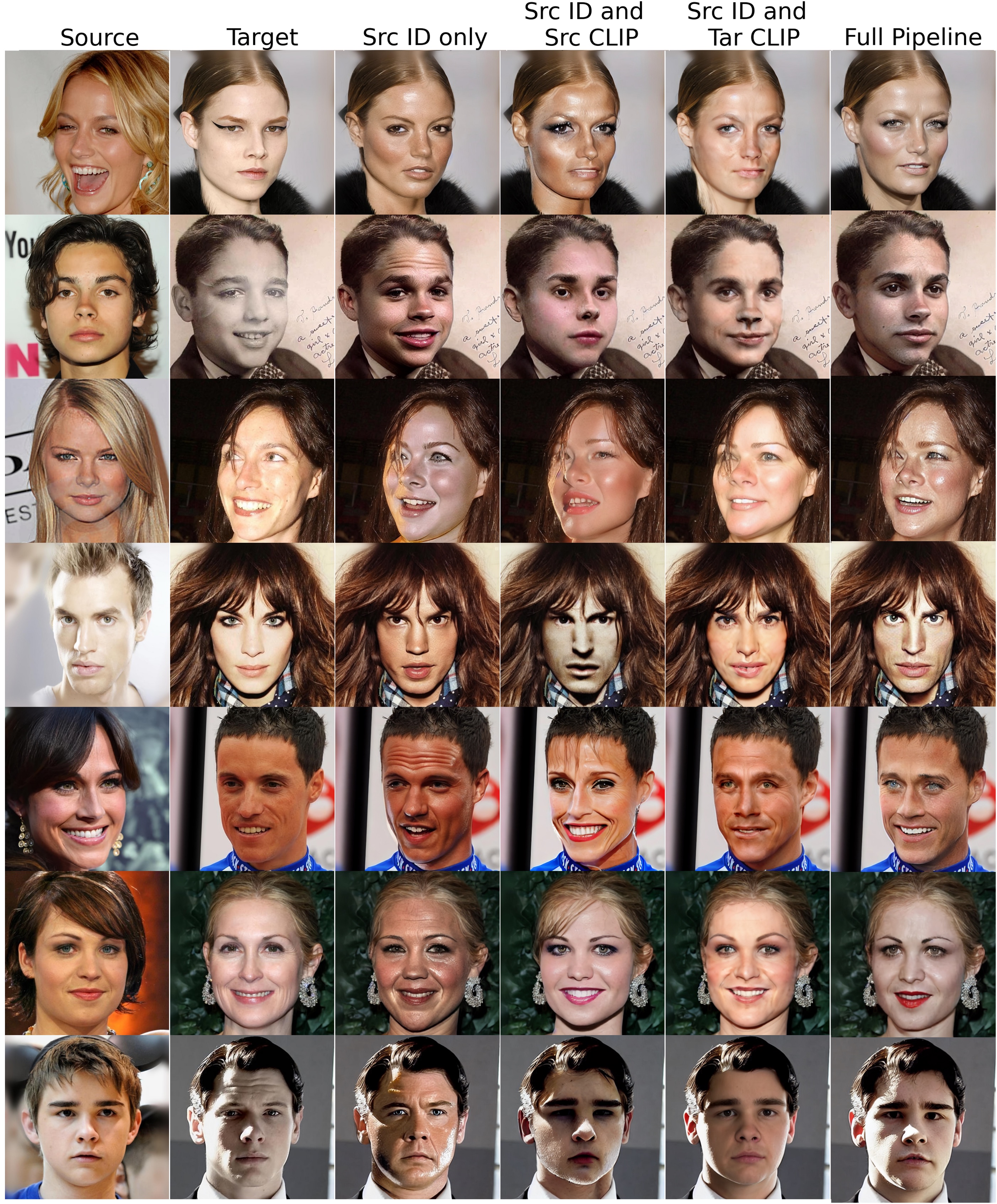}
        \caption{Ablation study on adding CLIP feature disentanglement.}
        \label{fig:clip_feature_img}
    \end{subfigure}
    \begin{subfigure}[b]{0.3465\linewidth}
        \centering
        \includegraphics[width=\linewidth]{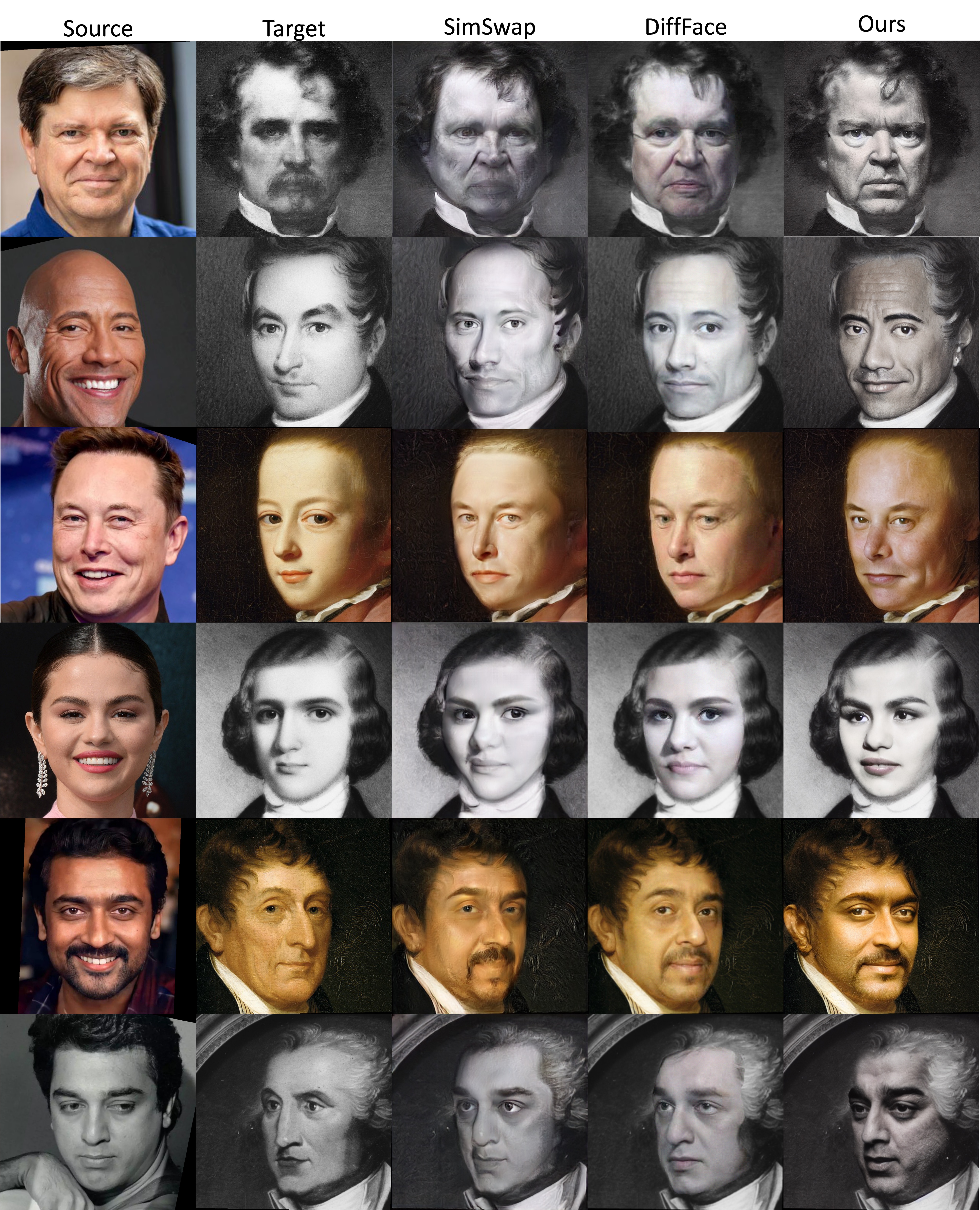}
        \caption{Out-of-distribution generalizability of our method. Target images are from \cite{rossler2019faceforensics++}.}
        \label{fig:out_of_dist}
    \end{subfigure}
    \caption{(a) Head swap results (b) Ablation on CLIP feature disentanglement (c) Qualitative results on out-of-distribution}
    \label{fig:comparison_results}
    \vspace{-1em}
\end{figure*}



\noindent\textbf{Evaluation protocol:} We obtain 1000 source and 1000 target images from the dataset and produce the 1000 swapped images.  We compute the FID \cite{heusel2017gans} between the swapped images. For the pose and expression, we use HopeNet\cite{doosti2020hope} and Deep3DFaceRecon \cite{deng2019accurate} respectively, and compare the target image and swapped image with L2 distance. To compute the ID retrieval we masked out the background of both the swapped image and source image, as that will contain different and unnecessary information. Then we extracted the ID features from ArcFace\cite{deng2019arcface}, and for each swapped face image, we conducted face retrieval by searching for the most similar faces among all the source faces, measured by cosine similarity, and then obtained the Top-1 accuracy and Top-5 accuracy. All results reported in this paper are obtained from a common benchmark that we implemented based on the prior works \cite{zhao2023diffswap,liu2023fine} benchmark system.




\noindent\textbf{Quantitative results:} We conduct experiments on
CelebA\cite{lee2020maskgan} dataset and compare our result with previous
methods in Table \ref{tab:CelebA}. Moreover, for a more comprehensive quantitative comparison, in Table \ref{tab:FFHQ}, we compare with the latest methods on the dataset FFHQ. Both in Table \ref{tab:CelebA} and Table \ref{tab:FFHQ}, the results demonstrate that our method surpasses the performance of previous methods in ID Retrieval and FID, indicating our ability to generate highly accurate swapped faces while better preserving the original source identity. 

\begin{table}[!htb]
\centering
\setlength{\tabcolsep}{2pt}
\begin{tabular}{lccccc}
\toprule
\multirow{2}{*}{\textbf{Method}}  & \multirow{2}{*}{\textbf{FID}$\downarrow$}    &   \multicolumn{2}{c}{\textbf{ID retrieval $\uparrow$}}   & \multirow{2}{*}{\textbf{Pose}$\downarrow$}   &  \multirow{2}{*}{\textbf{Expr.}$\downarrow$}    \\ 
 &&   Top-1 &  Top-5 \\
\midrule
MegaFS \cite{zhu2021one} &20.3 & 73.9\% & 80.3\% & 5.83 & 1.20  \\
HifiFace  \cite{wang2021hififace} &12.68 & 88.3\% & 94.3\% & 2.92 & 1.09 \\
SimSwap\cite{chen2020simswap} &10.7 & \underline{95.3}\% & \underline{98.6}\% & \underline{2.89} & 0.99  \\
E4S\cite{liu2023fine} &14.58 & 83.3\% & 90.8\% & 4.15 & 1.16  \\
FaceDancer\cite{rosberg2023facedancer} &12.83 & 0.10\% & 0.50\% & \textbf{2.15} & \textbf{0.70}  \\
DiffFace \cite{kim2022diffface} &10.29 & 94.6\% & 97.9\% & 36.1 & 2.15  \\
DiffSwap\cite{zhao2023diffswap} & \underline{9.16} & 75.6 \% & 90.9 \% & 2.48 & \underline{0.77} \\
 \midrule
\textbf{Ours}&\textbf{6.09 }& \textbf{98.8\%} & \textbf{99.6\%} & 3.51 & 0.96  \\
\bottomrule
\end{tabular}
\vspace{-1em}
\caption{Comparison on CelebA dataset. Best is in bold and second best is underlined.} 
\vspace{-2em}
\label{tab:CelebA}
\end{table}

\begin{table}[!htb]
\centering

\setlength{\tabcolsep}{2pt}
\begin{tabular}{lccccc}
\toprule
\multirow{2}{*}{\textbf{Method}}  & \multirow{2}{*}{\textbf{FID}$\downarrow$}    &   \multicolumn{2}{c}{\textbf{ID retrieval $\uparrow$}}   & \multirow{2}{*}{\textbf{Pose}$\downarrow$}   &  \multirow{2}{*}{\textbf{Expr.}$\downarrow$}    \\ 
 &&   Top-1 &  Top-5 \\
\midrule
MegaFS \cite{zhu2021one} &12.0 & 59.6\% & 74.1\% & 3.33 & 1.11  \\
HifiFace  \cite{wang2021hififace} &11.58 & 75.3\% & 87.1\% & 3.28 & 1.41 \\
SimSwap\cite{chen2020simswap} &13.8 & \underline{90.6\%} & \underline{96.4\%} & 2.98 & 1.07  \\
E4S\cite{liu2023fine} &12.38 & 70.2\% & 82.73\% & 4.50 & 1.31  \\
FaceDancer\cite{rosberg2023facedancer} &18.34 & 0.20\% & 0.90\% & \textbf{2.6} & \textbf{0.96}  \\
DiffFace \cite{kim2022diffface} & 8.59 & 87.2\% & 94.4\% & 3.80 & 2.28  \\
DiffSwap\cite{zhao2023diffswap} & \underline{8.58} & 78.2 \% & 93.6 \% & \underline{2.92} & 1.10 \\
 \midrule
\textbf{Ours}&\textbf{5.53}& \textbf{95.4\%} & \textbf{98.7\%} & 3.74 & \underline{1.04}  \\
\bottomrule
\end{tabular}
\vspace{-1em}
\caption{Comparison on FFHQ dataset.} 
\vspace{-1em}
\label{tab:FFHQ}
\end{table}


\noindent \textbf{HeadSwap:} HeadSwap has been previously explored by HeSer \cite{shu2022few} and HS Diffusion \cite{wang2023hs} primarily on plane background. Benefiting from the design of our mask shuffling, our method can also be extended to head swapping task with more challenging backgrounds. Further, unlike HS-Diffusion \cite{wang2023hs}, our method can well preserve the target's pose. Unfortunately, we are unable to provide comparison with their methods as they did not release their codes publicly. For our method, as displayed in Fig. \ref{fig:comparison_results} (a), with mask shuffling, especially in the hair region, the generated hairstyles are more realistic and properly aligned with the target compared to those generated without mask shuffling.

\noindent \textbf{Out-of-distribution generalizability for face-swapping:}
In Fig.~ \ref{fig:out_of_dist}, we showcase our model's adeptness at generating realistic face-swapped images on out-of-distribution images. We select target images from the manipulated FaceForensics++ dataset \cite{rossler2019faceforensics++} and compare with highly competitive methods, SimSwap\cite{chen2020simswap} and DiffFace\cite{kim2022diffface}. 

\begin{figure}[!htp]
    \centering
    \includegraphics[width=\linewidth]{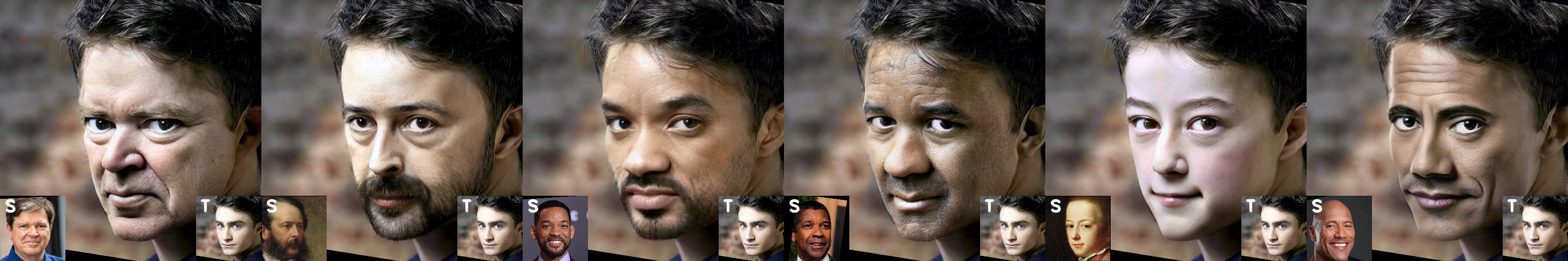}
    \vspace{-2em}
    \caption{Same Target different sources with skin color and texture variations. Zoom-in.}
    \vspace{-2em}
\label{fig:same_target_diff_source}
\end{figure}

\begin{figure}[!htp]
    \centering
    \includegraphics[width=\linewidth]{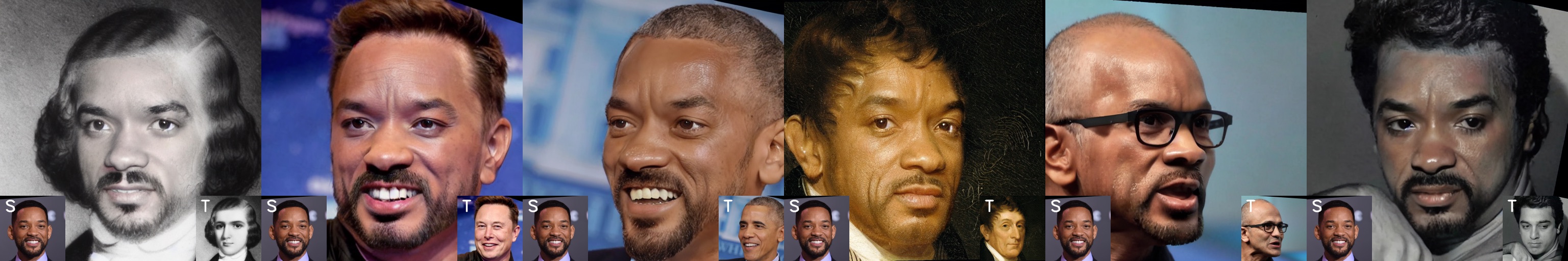}
    \vspace{-2em}
    \caption{Same Source different targets with variations in pose, expression, lighting, and texture. Zoom-in}
    \vspace{-2em}
\label{fig:same_source_diff_target}
\end{figure}

\noindent \textbf{Ablation study:} We analyze the role of CLIP feature disentanglement technique (Fig \ref{fig:comparison_results} (b)). In the absence of this module, the pose, expression and lighting of the swapped face result are poorly represented. Hence, It can illustrate the collaboration between this module and the ID decoupling module to make our face swap more superior. Further, we analyze how same source images can interact with multiple target images and vice-versa with different variations in skin tone and pose (Fig.~\ref{fig:same_source_diff_target} \& Fig.~\ref{fig:same_target_diff_source}). Regardless of different variations, our method consistently produces desirable swapping with high ID transferability.

\begin{table}[!htb]
    \centering
    \scalebox{0.9}{
    \begin{tabular}{l|c|c|c}
    \toprule
    \multirow{2}{*}{\textbf{Method}} & \textbf{Per image} & \multirow{2}{*}{\textbf{Resolution}} & \textbf{Train Time} \\
    & \textbf{inference} & & \textbf{GPU-days} \\
    \midrule
    DiffFace \cite{kim2022diffface} & 32 s & 256 & 80 \\
    DiffSwap \cite{zhao2023diffswap} & 9.4 s & 256 & unknown \\
    Ours & \textbf{4.7 s} & \textbf{512} & \textbf{18} \\
    \bottomrule
    \end{tabular}}
    \vspace{-1em}
    \caption{Comparison of inference time and  training cost.} 
    \label{tab:time}
    \vspace{-1em}
\end{table}



\noindent \textbf{Resource consumption:} Although diffusion models have strong generative capabilities, the simultaneous multi-step denoising is accompanied by strong compute resources and time consumption \cite{song2022denoising}. Compared to DiffFace and DiffSwap, ours is more efficient in inference time, training cost, and can produce high res. output (Table \ref{tab:time}). For diffswap, we cannot estimate the train time. However, they use 8 A100s with global batch size 32 for 100k iters, compared to 2 A100s with global batch size of 4 for 140k iters on ours. Although we present results and analysis for 50-step inference in this manuscript, our method is capable of producing good swapped images with just 5 steps. See the appendix.\ref{sec:effect on number of steps} for more analysis on this. Additionally, refer to the appendix for more qualitative results and other details. 





\section{Conclusion \& Limitations}

We proposed a train-time diffusion-based inpainting pipeline for face-swapping to obtain realistic swaps. Our introduction of a disentangled CLIP feature further improves the pose and expression perseverance. Furthermore, we propose a simple mask shuffling technique to even handle headswapping task. While our method significantly boosts both the performance (in qualitative and quantitative results) and efficiency (i.e. inference time and training cost), there is still room for improvement, especially under extreme pose and expression variations which we leave for future work.
{\small
\bibliographystyle{ieee_fullname}
\bibliography{egbib}
}
\clearpage


\appendix
\section{Appendix}
\begin{figure*}[!htb]
    \centering
    \includegraphics[width=\linewidth]{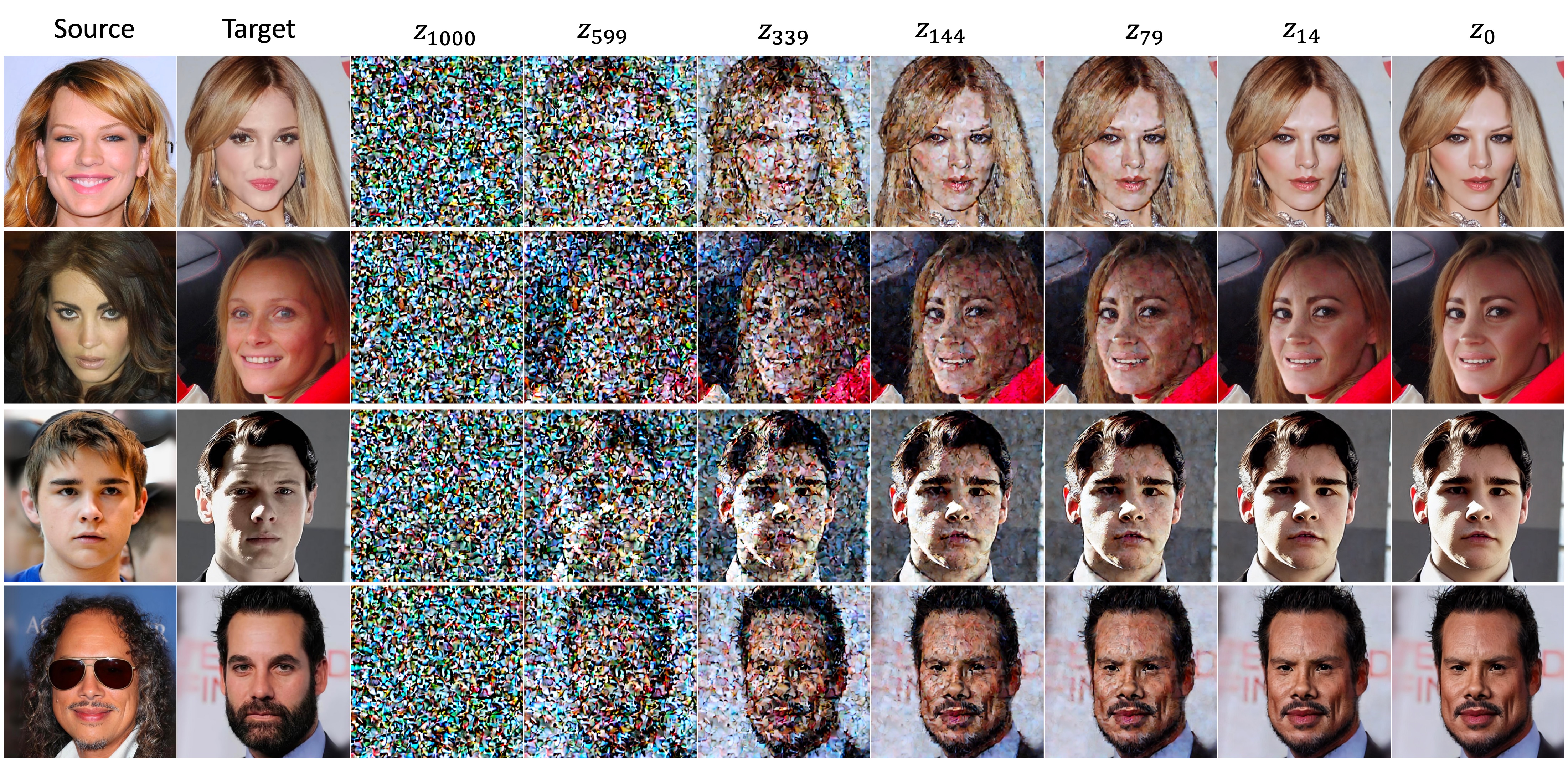}
    \vspace{-1em}
    \caption{Denoising process visualization. The images shows the decoded output of noisy latents ($\mathcal{D}(z_t)$) through DDIM process.}
    \label{fig:diffusion_process}
\end{figure*}

\begin{figure*}[!htb]
    \centering
    \includegraphics[width=\linewidth]{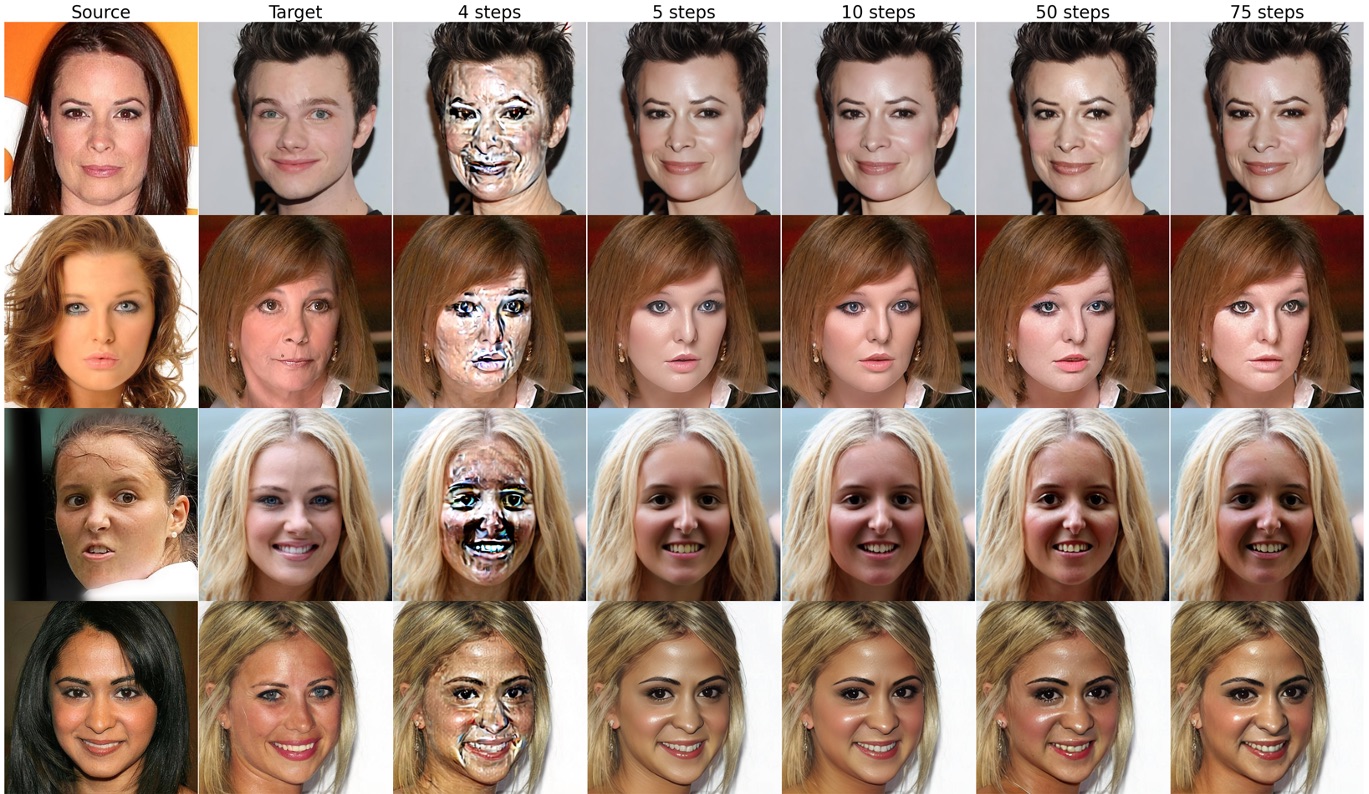}
    \caption{Comparison for total number of denoising steps using DDIM with our model.}
    \label{fig:DDIM_steps}
\end{figure*}

\begin{figure*}[!htb]
    \centering
    \includegraphics[width=\linewidth]{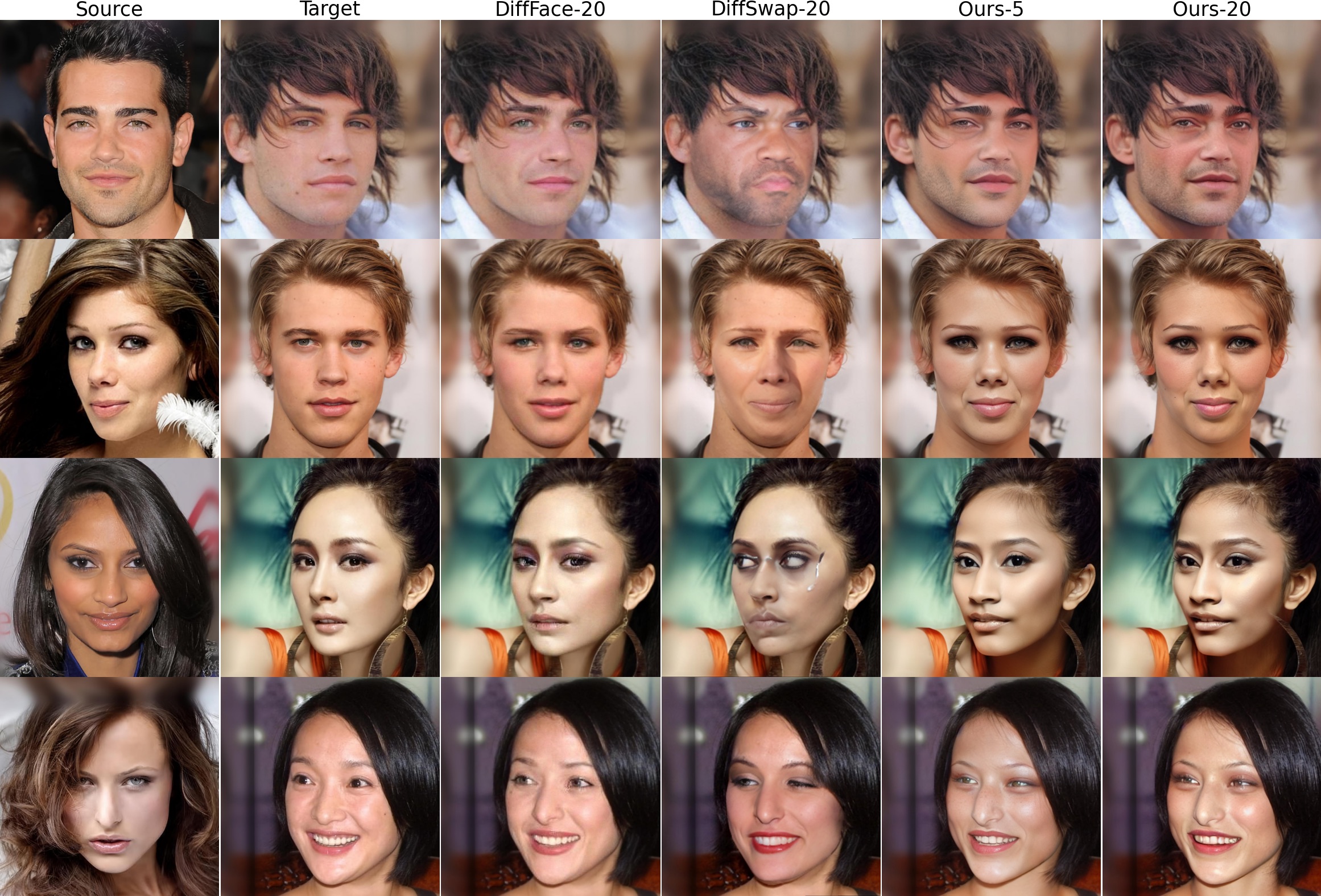}
    \caption{Comparison of the effect of total number of denoising steps using DDIM with other approaches. While both DiffSwap\cite{zhao2023diffswap} and DiffFace\cite{kim2022diffface} are failing to transfer identity and making artifacts, our method produces superior swapped images even with 5 steps}
    \label{fig:DDIM_steps_others}
\end{figure*}

\subsection{Diffusion Process Visualization}
\label{sec: diff_process_visualize}

Fig \ref{fig:diffusion_process} shows the source, target, and the decoded output images through the diffusion process. Here, we show the denoising through 75 DDIM steps. The $z_{1000} \cdots z_0$ images in Fig \ref{fig:diffusion_process} correspond to $0^{th}$, $30^{th}$, $50^{th}$, $65^{th}$, $70^{th}$, $73^{th}$, and $75^{th}$ DDIM step respectively. We observe initial steps recover the basic structure of the swapped image while a few later ($65^{th} \text{ to } 75^{th}$) steps refine the image. 


\subsection{Effect on Number of Steps in DDIM}
\label{sec:effect on number of steps}

Existing diffusion-based works on face-swapping use time-consuming denoising steps. DiffSwap\cite{zhao2023diffswap} uses 200 steps with masked fusion at inference. DiffFace\cite{kim2022diffface} uses 75 steps which consumes a huge amount of time (approximately 9 hours to perform 1000 swaps) due to gradient computation in their inference strategy. We use 50 steps to compare the inference time in which we showed our method performs a magnitude faster inference than DiffFace\cite{kim2022diffface} and roughly twice faster than DiffSwap \cite{zhao2023diffswap}. Further, all the results we show in the main manuscript use 50 steps. However, to analyze the effect of the number of DDIM steps with our algorithm, we show further analysis of qualitative images with varying numbers of DDIM steps in Fig \ref{fig:DDIM_steps}.

In contrast to existing methods that rely on complex inference processes for face-swapping, often failing to produce satisfactory results and fail to transfer identity features well with minimal denoising steps, our approach stands out by achieving superior swapping outcomes even with as few as 5 steps (see Fig.~\ref{fig:DDIM_steps_others}), which significantly reduces the computational overhead, slashing the inference time to approximately one-tenth of the duration required for 50 steps.





\subsection{Additional Implementation Details}
\label{sec:additional_implementaion}

Building upon the implementation details outlined in the main manuscript, we provide additional specifications for clarity. We adopt a pre-trained stable diffusion checkpoint, akin to the framework introduced in \cite{yang2022paint}, with a modification involving 9 channels. We use AdamW optimizer with learning rate $1e-5$ and other default parameters. Latent size is $64 \times 64$. The condition feature dimension $D$ is 768. In condition generation, the CLIP weight $w_{clip}$, ID feature weight $w_{id}$, and the landmark feature weight $w_{lm}$ are 1.0, 10.0, and 0.05 respectively. The number of DDIM steps in our training pipeline $N=4$. The output image resolution is $512 \times 512$.

\subsection{Additional Information on Face shape Augmentation}
\label{sec: additional info Face shape augmentation}

To facilitate face shape augmentation, an image elastic deformation approach \cite{he2023shift,lin2024boosting} based on Thin Plate Spline (TPS) transformation \cite{bookstein1989principal} was employed. Specifically, we first generate a 2D grid of points of the same size as the face mask. Then, we set up a set of control net points $O$ on the grid. Next, we add random noise $\delta$ to control net points $O$ and obtain $P$. The intensity of the noise is controlled by a scaling factor $s$ to enable precise modulation. By utilizing the two sets of control points, we can obtain an interpolation function that acts on the entire mask, allowing us to achieve our mask augmentation while ensuring coherence and consistency in subsequent transformations. This is crucial for enhancing diversity and realism in augmented face shapes, providing smooth and continuous deformation while preserving structural integrity. We use this face shape augmentation to get the inpaint image in our pipeline. We use a random scale $s$ sampled uniformly from the range 0.5 to 1.

\subsection{Additional Qualitative Results}

\begin{figure*}[!htb]
    \centering
    \includegraphics[width=0.9\linewidth]{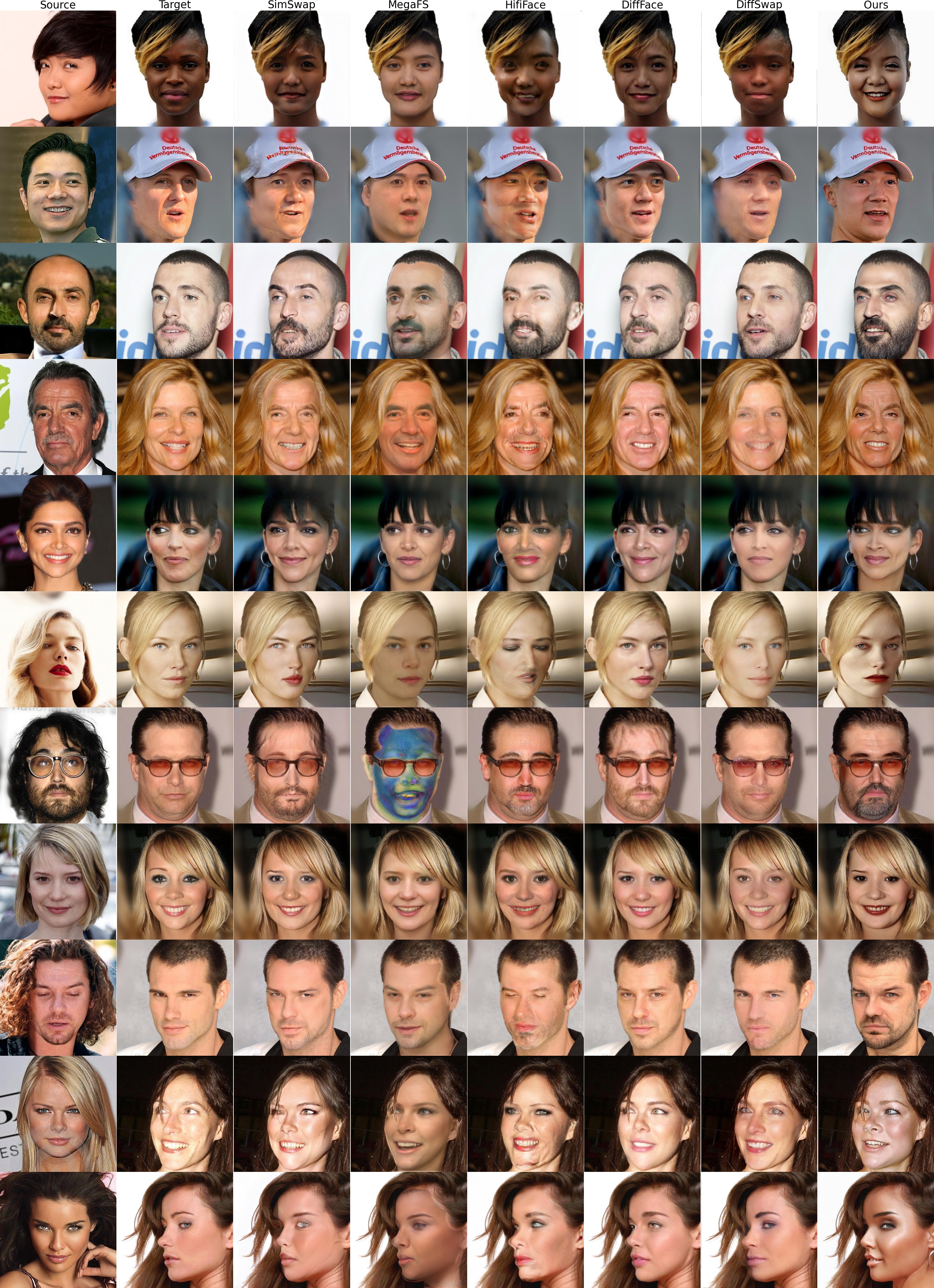}
    \caption{Qualitative comparison on CelebA dataset. Better viewed in Zoom}
    \label{fig:celeb_supp}
\end{figure*}

\begin{figure*}[!htb]
    \centering
    \includegraphics[width=0.9\linewidth]{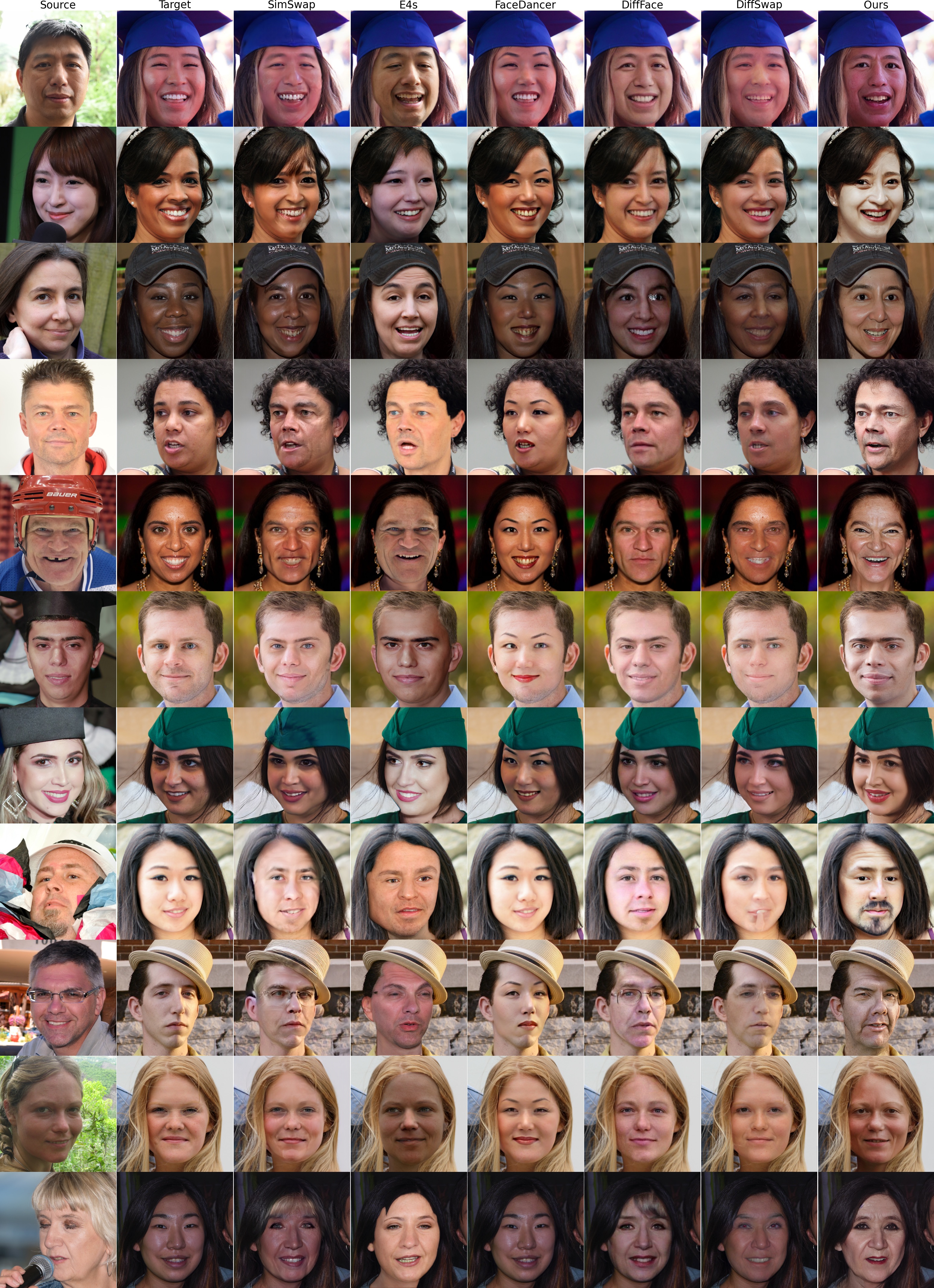}
    \caption{Qualitative comparison on FFHQ dataset. Better viewed in Zoom}
    \label{fig:FFHQ_supp}
\end{figure*}

We provide more qualitative comparison for Face Swapping on CelebA dataset (Fig \ref{fig:celeb_supp}) and FFHQ dataset (Fig \ref{fig:FFHQ_supp}). In the examples, we observe, our method produces smoother boundaries and photo-realistic images. Unlike other works which suffer in merging the swapped image and the target's hair and background, which often results in a visible merging boundary our method seamlessly blends the swapped image as there is no separate step of merging. Moreover, other works produce a lot of artifacts, especially in challenging situations such as extreme pose variations (E.g., last row of Fig \ref{fig:celeb_supp}), and occlusions or accessories in source (E.g., third last row of Fig \ref{fig:FFHQ_supp}). 

\begin{figure*}[!htb]
    \centering
    \includegraphics[width=0.45\linewidth]{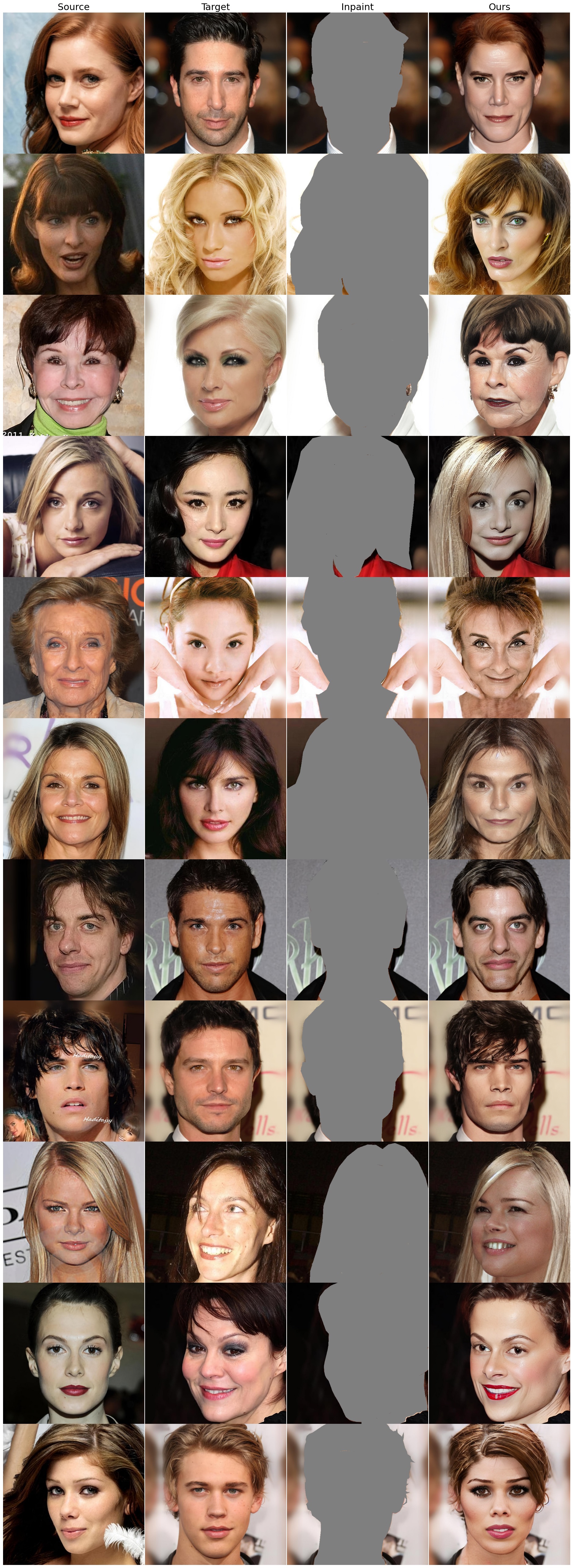}
    \caption{Additional Head Swap outcomes. Better viewed in Zoom}
    \label{fig:Head_swap_supp}
\end{figure*}

Further, we provide additional head-swapping qualitative images in Fig \ref{fig:Head_swap_supp}. Despite challenging masks, our approach is capable of producing realistic head swaps while preserving the target pose and expression. 

\subsection{Societal Impact} With the advancements in deep learning, creating swapped face has become easier and social media platforms have made it easier for them to spread rapidly. Everything has two sides. If facial swapping technology is used for the advancement of productivity, such as in movie scenes, it can greatly enhance productivity. However, if this technology is exploited by malicious individuals, face swapping may pose a significant threat to society. We are committed to developing powerful face swapping technologies that have a beneficial impact on society. The purpose of our research is also to promote the healthy development of this technology. Furthermore, we control the generation of vulnerable images in our method's safety check via Stable Diffusion \cite{rombach2022high}.

\end{document}